\documentclass{article}
\usepackage[pdftex]{graphicx,color}
\usepackage[fleqn]{amsmath}
\usepackage{latexsym}
\usepackage{amssymb}
\usepackage{subcaption}
\usepackage{authblk}
\usepackage{bm}

\title{Clustering and Data Augmentation to Improve Accuracy of Sleep Assessment and Sleep Individuality Analysis}
\author[1]{Shintaro Tamai}
\author[2,3]{Masayuki Numao}
\author[2]{Ken-ichi Fukui}
\affil[1]{Graduate School of Information Science and Technology, Osaka University, Japan}
\affil[2]{SANKEN, Osaka University, Japan}
\affil[3]{Kyoto Tachibana University, Japan}

\begin{document}
\maketitle
\begin{abstract}
Recently, growing health awareness, novel methods allow individuals to monitor sleep at home. Utilizing sleep sounds offers advantages over conventional methods like smartwatches, being non-intrusive, and capable of detecting various physiological activities. This study aims to construct a machine learning-based sleep assessment model providing evidence-based assessments, such as poor sleep due to frequent movement during sleep onset. Extracting sleep sound events, deriving latent representations using VAE, clustering with GMM, and training LSTM for subjective sleep assessment achieved a high accuracy of 94.8\% in distinguishing sleep satisfaction. Moreover, TimeSHAP revealed differences in impactful sound event types and timings for different individuals.
\end{abstract}

\section{Introduction}
Sleep plays an extremely important role in human health. Ensuring an adequate amount of high-quality sleep is essential for maintaining physical health and psychological balance. Professional measurement of sleep state is mainly conducted through Polysomnography (PSG) \cite{PSG}. However, PSG involves a significant physical burden on the subjects and is difficult to measure without specialized facilities or hospitals. In recent years, evaluation methods utilizing wearable devices have been developed with the aim of facilitating sleep assessment \cite{Kwon}. However, the information that can be obtained through a smartwatch is limited, typically encompassing data such as acceleration and heart rate. While EEG-based sleep monitoring offers high accuracy, the requirement to wear headgear, even for a single-channel EEG headset \cite{singleEEG}, presents a significant burden.

Unlike EEG \cite{SleepEEGNet} or pressure sensors \cite{Bed}, sound-based method is non-contact and easily collectible. Sleep sounds refer to the sounds related to biological activities such as snoring, body movements, coughing, and environmental noises emitted during sleep. Methods using sleep sounds have advantages over conventional methods, such as being non-contact and capable of detecting many biological activities. Traditionally, sleep evaluation based on sleep sounds has primarily focused on the detection of sleep apnea syndrome\cite{nakano,luo}, and there is still limited research regarding the evaluation of sleep quality.

In existing deep learning-based sleep quality estimation using sleep sounds \cite{ishi,chen}, the basis of evaluation was a black box. Therefore, in this study, we propose a sleep quality classification model based on machine learning using sleep sounds that can provide rationales such as ``poor sleep due to frequent tossing and turning during sleep''. By providing rationales, it may be possible to contribute to improving the user's sleep quality.

In the previous studies \cite{ishi,chen}, the feature representation of sleep sound events was obtained using Variational AutoEncoder (VAE) \cite{VAE}, and the classification of sleep quality was based on Long Short-Term Memory (LSTM) \cite{LSTM}.
In this study, we extended a classification method based on VAE-LSTM by adding explainability to the classification of sleep quality. To provide explainability, it is necessary to identify the types of sound events; however, labeling each event individually is challenging. Therefore, we introduced sound event clustering to facilitate the labeling of sound events. Additionally, we proposed a method to obtain explanations for the trained classification model using TimeSHAP\cite{timeshap}. This method can present the contribution of each sound event cluster to the classification. Through an experiment with three subjects conducted in a home environment over approximately 20 days, we demonstrated that it is possible to analyze which types of events are involved in determining sleep quality and to identify the unique sleep characteristics of each subject. 

Moreover, while conventional methods used discrete power spectra of sound events as feature vectors for input to the VAE, in this study, we proposed a method that uses the probability of belonging to each cluster as input. By combining this with data augmentation, an improvement trend in classification accuracy was observed in experiments with the three subjects.

The novelty of the proposed method compared to the conventional method of sleep quality classification using VAE-LSTM is as follows:
\begin{itemize}
\item We proposed a framework that introduces clustering of sleep sound events, making it possible to evaluate the types of sound events that affect sleep quality classification by LSTM.
\item We proposed a method that represents sleep sound events using the probability of belonging to each cluster. Through experiments, we demonstrated that combining this with data augmentation could lead to improved accuracy in sleep quality classification.
\end{itemize}

\section{Related Works}
Wu \textit{et al.} \cite{wu} proposed a method based on Kernel Self-Organizing Map to visualize patterns of sleep personalities. In this study, they confirmed the relationship between sleep stages and sleep sounds through comparison with PSG data, demonstrating the validity of sleep evaluation using sleep sounds. Here, sleep stages refer to the depth of sleep (such as REM sleep or non-REM sleep) corresponding to different states of sleep. Furthermore, Wu \textit{et al.} \cite{wu2} trained a time-series model of sleep patterns using Hidden Markov Model (HMM) based on sleep sound data recorded at home and performed sleep quality classification using Support Vector Machine (SVM).

Ishimaru \textit{et al.} \cite{ishi} proposed a sleep evaluation method considering individual differences using a combination of Variational Domain Adversarial Neural Network with VAE and Domain Adaptation, and LSTM based on sleep sound data recorded at home. Chen \textit{et al.} \cite{chen} performed sleep quality classification considering not only sleep sounds but also external factors such as light and humidity, as well as internal factors such as BMI and health status, using VAE and LSTM. Additionally, in this study, they examined the impact of each factor on sleep by weighting external and internal factors with Gated Residual Network.

These studies \cite{ishi,chen} had a common issue where the basis of evaluation was considered a black box, making it difficult to understand the extent to which kind of sleep sound event influences the evaluation. In Chen \textit{et al.}'s study \cite{chen}, although they investigated the impact of external and internal factors on sleep, they were unable to evaluate the influence of each sleep sound event. Additionally, Wu \textit{et al.} \cite{wu2} revealed significant differences in transition probabilities in HMM depending on the quality of sleep, demonstrating the basis for sleep quality classification, but they did not specifically identify important sleep sound events.

In this study, we address this issue by clustering sleep sound events and using TimeSHAP to analyze the SHAP values \cite{shap} of each cluster. This allows us to interpret the interpretation of sleep quality classification for a night's sleep and identify important sleep sound events and individual-specific features, thus resolving this problem.

Another issue arises in the targets of sleep status estimation. Many studies focus on estimating a sequence of ``sleep stages'' throughout the night \cite{SleepEEGNet,Bed,Radio,Phan}, including based on sound \cite{Tran,Dafna}. Sleep stages serve as the gold standard for sleep research and the diagnosis of sleep disorders in clinical settings. However, we believe that sleep stages may not be an adequate metric for daily-life sleep monitoring, especially when presented to non-experts. The depth of sleep does not necessarily directly reflect the quality of sleep. As there is no widely accepted common definition of sleep quality \cite{SleepQuality}, we propose relying on subjective sleep satisfaction as a metric for daily-life monitoring, rather than sleep stages, or WASSO (Wakefulness After Sleep Onset) and Sleep Efficiency, which are induced by sleep stages. 

\section{The Proposed Method}
\subsection{Overview}
\begin{figure*}[h]
    \begin{center}
        \includegraphics[width=1\textwidth]{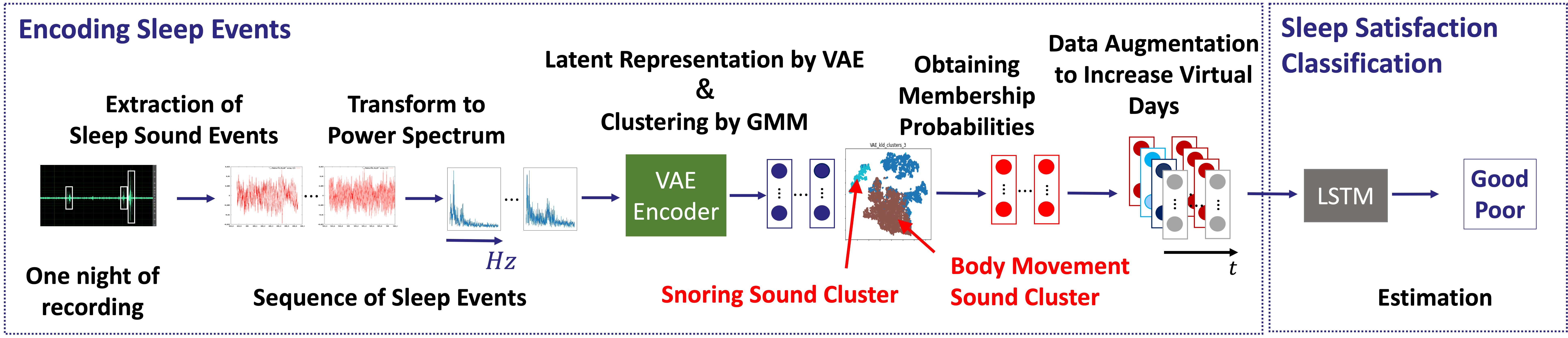}
        \caption{Overview of the proposed sleep satisfaction classification introducing clustering of sound events}
        \label{fig:g}
    \end{center}
\end{figure*}

In this study, building upon existing research \cite{ishi,chen}, we introduce a novel approach by incorporating clustering of sleep sound events to propose a highly accurate and interpretable method for sleep quality classification. Clustering of sleep sound events facilitates the interpretation of each event, enabling the identification of important events for sleep quality classification. By clustering sleep sound events, it becomes easier to assign meaning to each event, allowing for the identification of crucial events for sleep quality classification.

The following is an overview of the proposed method as illustrated in Figure \ref{fig:g}. First, sleep sound events are extracted from overnight recordings, and then transformed into the frequency domain using Fast Fourier Transform (FFT). Next, the power spectrum of these sleep sound events serves as input to obtain latent representations through Variational AutoEncoder (VAE). Subsequently, the latent representations of these sleep sound events are clustered using Gaussian Mixture Model (GMM), and the membership probabilities to each cluster are determined. Data augmentation is performed by sampling events multiple times per night to simulate an increase in the number of nights. Then, the probabilities of belonging to each cluster, data augmentation, are fed into an LSTM for sleep quality classification. Finally, TimeSHAP is applied to the trained LSTM model to analyze the SHAP values of each cluster, allowing for the interpretation of important sleep sound events and individual-specific features for sleep quality classification throughout a night.

\subsection{Preprocessing}
First, sleep sound events are extracted from the audio recorded continuously overnight. Similar to previous studies \cite{ishi,chen}, we employed Kleinberg's burst extraction method \cite{burst} to extract sleep sound events. The burst extraction method is based on the assumption that the amplitude of the waveform follows a normal distribution. It identifies segments that are estimated to be generated persistently from a normal distribution with a larger variance compared to stationary noise.

The extracted sounds are transformed into the frequency domain by applying the Fast Fourier Transform (FFT), and the power spectrum is used as the input vector for the Variational AutoEncoder (VAE). Here, the input vector consists of a 2,400-dimensional vector, with discrete points discretized at 10 Hz intervals ranging from 10 Hz to 24,000 Hz.

Furthermore, due to the characteristics of the burst extraction method, even continuous sounds may be fragmented and extracted separately. Therefore, similar to previous research \cite{ishi}, in this study, we performed subsampling of sleep sound events by setting a threshold on the time interval between the occurrences of sleep sound events.

\subsection{Sleep Sound Event Clustering}
The Variational AutoEncoder (VAE) learns a latent representation by training on the set of sleep sound events as input. Typically, the mean squared error (MSE) is used as the reconstruction error term in VAE. However, MSE treats each discretized point of the spectrum independently when measuring the error, neglecting the shape of the power spectrum and failing to capture subtle differences between similar sleep sound events \cite{ishi}. Therefore, similar to previous research \cite{ishi,chen}, in this study, we normalized the power spectrum so that its sum equals one, treating it as a probability distribution, and employed the Kullback-Leibler divergence (KLD) as the reconstruction error term in VAE. By measuring the error as a probability distribution, we aim to acquire latent representations that minimize the distance between similar spectrum in the latent space.

After training the VAE, we use the encoder to obtain the latent representation $\mathbf{z}$ of the sleep sound event $\mathbf{x}$. In this process, only the mean vector is used from the latent representation.
\begin{equation}
 \mathbf{z}= \bm{\mu}(\mathbf{x})=\mathrm{VAE}_{\mathrm{enc}}(\mathbf{x}).
\end{equation}

Next, we cluster the set of latent representations $\mathbf{z}$ of sleep sound events using Gaussian Mixture Model (GMM) and obtain the membership probabilities $\mathbf{p}$ for the clusters.

\subsection{Data Augmentation to Virtually Increasing the Number of Days}
Based on the membership probabilities $\mathbf{p}$ to each cluster obtained through clustering, we construct the event sequence data for a night as the input data for LSTM, denoted as $\mathbf{x_{inp}}=[\mathbf{p_1},\mathbf{p_2},\cdots,\mathbf{p_l}]$, where $l$ represents the number of events in a night. As the number of events in a night ranges from approximately 200 to 4,000, inputting the entire sequence of sleep sound events for a night would result in a sequence length that is too long for LSTM input.

\begin{figure}[h]
  \centering 
  \includegraphics[width=0.7\linewidth]{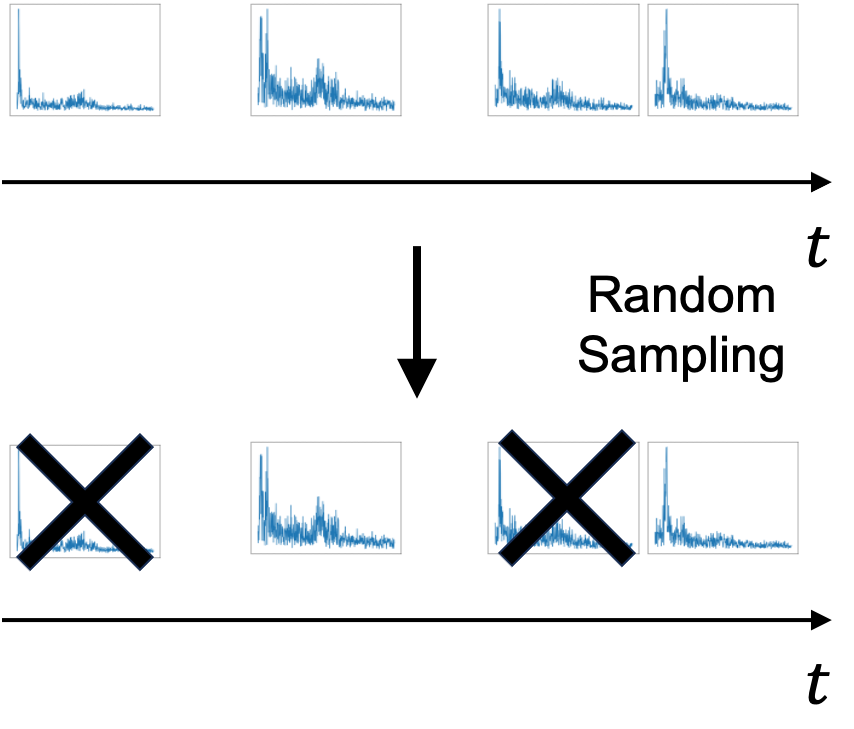}
  \caption{Random sampling of sound events for data augmentation}
  \label{fig:aug}
\end{figure}

Therefore, similar to previous studies \cite{ishi,chen}, we randomly sample the event sequences at the event level (as shown in Figure \ref{fig:aug}) and utilize them as inputs for LSTM. In this process, the order of event sequences is preserved. The number of sampling events is set individually for each subject. Furthermore, this study proposes data augmentation to virtually increase the number of days. By performing multiple random samplings, different event sequences are generated, effectively expanding the virtual number of days. This augmentation process is applied to all days.

\subsection{Sleep Satisfaction Classification by LSTM}
We train an LSTM for subjective evaluation estimation using a sequence-to-one approach to perform sleep quality classification. The subjective evaluation to be estimated is ``satisfaction'', which is a binary classification between ``satisfied'' and ``unsatisfied'', excluding ``neutral''. Here, sleep satisfaction significantly influences all aspects of subjective evaluations related to daytime activities \cite{lenn}. Therefore, considering the subjective satisfaction as the supervisory information representing sleep quality is reasonable.

\subsection{Interpretation of Sleep Satisfaction Classification by TimeSHAP}
Finally, we apply TimeSHAP to the trained LSTM to interpret the sleep quality classification for each level of satisfaction. TimeSHAP extends the SHapley Additive exPlanations (SHAP) method \cite{shap} to interpret sequences of events. By analyzing the SHAP values of each cluster and the interpretation of representative clusters, we can identify important sleep sound events, time periods, and individual characteristics. The interpretation of clusters was conducted by randomly selecting sound events, listening to the actual sounds, and estimating the types of events.

\section{Experiments}
\subsection{Dataset}
In this study, data collected with the cooperation of participants from a wide range of age groups was utilized. The data collection was conducted with the approval of the Ethics Committee of SANKEN, Osaka University. Participants were asked to record their sleep sounds in their private bedrooms at home for one month. Recording was done using a smartphone (Zenfone Live Android 7.0). The sampling frequency of the microphone is 48 kHz. Additionally, participants filled out questionnaires. Before bedtime, they provided answers regarding their physical and mental fatigue, presence of illnesses or injuries, etc. After waking up, they rated their sleep satisfaction and the indoor environment during sleep. Sleep satisfaction was rated on a five-point scale: ``very satisfied,'' ``satisfied,'' ``neutral,'' ``unsatisfied,'' and ``very unsatisfied.'' Moreover, participants' profiles such as age and gender were recorded.

In the experiments of this study, data from days without the use of air conditioning equipment such as air conditioners or fans, based on questionnaire responses to ensure minimal noise in sleep sounds, were selected. Additionally, data from participants who had a cold or injury were excluded. From the selected participants, three individuals were chosen based on having a high number of days with satisfied or unsatisfied sleep. Table \ref{table:hikensya} presents the profiles of the three selected participants and the number of days with satisfied and unsatisfied. 

\begin{table}[t]
 \caption{Subjects' basic information}
 \label{table:hikensya}
 \centering
  \begin{tabular}{c||cccccc}
   \hline
   subject ID & age & gender & \# satisfied days & \# unsatisfied days & total rec. time \\
   \hline \hline
   1 & 43& F& 8& 10 & 96h\\
   2 & 23& M& 12& 8 & 125.5h\\
   3 & 21& F& 10 & 11 & 147.5h\\
   \hline
  \end{tabular}
\end{table}

In prior studies \cite{ishi,chen}, a common model was trained on the participants group, whereas this study trains individual models for each person. Since each participant contributed data for approximately 20 days, we consider this to be a sufficient number of days for training individual models.

\subsection{Classification Result of Sleep Satisfaction}
In this experiment, we conducted sleep quality classification using the proposed method and examined its classification accuracy. First, we trained the LSTM with varying conditions, the dimensionality of VAE latent space, and the number of clusters in GMM. The LSTM had a single hidden layer with 50 nodes, and a dropout rate of 0.2 was applied between the hidden and output layers. The output layer utilized a Sigmoid activation function. We employed the Adam optimizer \cite{adam} with a learning rate of 0.001. And a binary cross-entropy is used for the loss function. The sampling size of the sound events used in LSTM training was set according to the number of events for each participant. The multiplying factor of the augmentation was set to 5,000 for all participants. To increase the number of trials, we performed 4-Fold Cross Validation five times with different random seeds to determine the optimal conditions for each participant. The training was terminated after 10 epochs, as the loss on the training data had generally converged. Table \ref{table:n_events} shows the number of sleep sound events used for the training data (before data augmentation) and test data for each subject.

\begin{table}[t]
\caption{The number of sleep sound events for each subject. It varies due to the different number of days in the cross-validation splits.}
  \centering
  \begin{tabular}{c|cc}
  \hline
   ID & Training data (w/o data augmentation) & Test data\\ \hline
   1 & 5,200 (13 days) / 5,600 (14 days) & 2,000 (5 days) / 1,600 (4 days)\\
   2 & 3,000 (15 days) & 1,000 (5 days)\\
   3 & 7,500 (15 days) / 8,000 (16 days) & 3,000 (6 days) / 2,500 (5 days)\\
   \hline
  \end{tabular}
  \label{table:n_events}
\end{table}

The results are presented in Tables \ref{table:D-S051}, \ref{table:N-S003}, and \ref{table:N-S026}. Hereafter, all “$\pm$” in the tables indicate the standard deviation. It was observed that the optimal conditions for the dimensionality of VAE latent space and the number of clusters for achieving high accuracy varied among participants.

\begin{table}[t]
\caption{The average of accuracy for different dimensionality of latent representation of VAE and the number of clusters (Subject 1)}
  \centering
  \begin{tabular}{c||ccc} 
  \hline
    \# clusters \textbackslash VAE dim. & 20 & 100 & 150\\ \hline\hline
    3 & 0.698$\pm$0.058& 0.705$\pm$0.056 &0.640$\pm$0.058 \\
    6 & \textbf{0.892} $\pm$ \bf{0.036} &0.818$\pm$0.058 &0.813$\pm$0.080\\
    9 & 0.782$\pm$0.070& 0.748$\pm$0.056 &0.838$\pm$0.076\\ \hline
  \end{tabular}
  \label{table:D-S051}
\end{table}

\begin{table}[t]
\caption{The average of accuracy for different dimensionality of latent representation of VAE and the number of clusters (Subject 2)}
  \centering
  \begin{tabular}{c||ccc} 
  \hline
    \# clusters \textbackslash VAE dim. & 20 & 100 & 150\\ \hline\hline
    3 & 0.710$\pm$0.066&\textbf{0.826$\pm$0.061}&0.640$\pm$0.037 \\
    6 & 0.763$\pm$0.085&0.740$\pm$0.097&0.782$\pm$0.091\\
    9 & 0.740$\pm$0.020&0.720$\pm$0.081&0.690$\pm$0.058\\ \hline
  \end{tabular}
  \label{table:N-S003}
\end{table}

\begin{table}[t]
\caption{The average of accuracy for different dimensionality of latent representation of VAE and the number of clusters (Subject 3)}
  \centering
  \begin{tabular}{c||ccc} 
  \hline
    \# clusters \textbackslash VAE dim. & 20 & 100 & 150\\ \hline\hline
    3 & 0.772$\pm$0.040&0.732$\pm$0.095&0.783$\pm$0.096 \\
    6 & 0.862$\pm$0.060&0.795$\pm$0.042&\textbf{0.948$\pm$0.050}\\
    9 & 0.857$\pm$0.070&0.895$\pm$0.064&0.842$\pm$0.071\\ \hline
  \end{tabular}
  \label{table:N-S026}
\end{table}

Next, we investigated the effectiveness of the proposed data augmentation and the impact of multiplying factor on accuracy. The multiplying factor ranged from 200 to 5,000, and we compared the obtained accuracies at each factor. The dimensionality of the VAE latent space and the number of clusters for each participant were set to the optimal parameters determined in the aforementioned experiments. The results are presented in Table \ref{seido}. It is evident from the results that the accuracy significantly improved with data augmentation for all participants. Thus, it can be concluded that the proposed method enables accurate classification of sleep quality for a night. Moreover, it is observed that higher multiplying factors tend to result in higher accuracy compared to lower factors.

\begin{table}[t]
\caption{The average accuracy for different magnification of data augmentation}
  \centering
  \begin{tabular}{c||ccc} 
  \hline
    multiplying factor \textbackslash subject ID & 1 & 2 & 3\\ \hline\hline
    None & 0.800 $\pm$ 0.044&0.600$\pm$0.00&0.717$\pm$0.066 \\
    200 & 0.845$\pm$0.075&0.740$\pm$0.102&0.858$\pm$0.068\\
    500 & 0.818$\pm$0.053&0.730$\pm$0.129&0.827$\pm$0.076\\
    1,000 & 0.845$\pm$0.075&0.780$\pm$0.081&0.867$\pm$0.100\\
    2,000 & \textbf{0.903$\pm$0.040}&\textbf{0.840$\pm$0.073}&0.893$\pm$0.048\\
    5,000 & 0.893$\pm$0.036&0.826$\pm$0.061&\textbf{0.948$\pm$0.050}\\ \hline
  \end{tabular}
  \label{seido}
\end{table}

Here, for subjects 1 and 2, performance peaked at 2,000x data augmentation. In this data augmentation process, increasing the augmentation rate did not always lead to improved performance. While the accuracy on the training data was between 0.88 and 0.91 at 2,000x augmentation, it increased to between 0.96 and 0.98 at 5,000x. However, since this data augmentation method generates new event sequences by sampling sound event sequences, excessive data augmentation may lead to overfitting.

For each participant, the highest accuracy achieved was 90.3\% for subject 1, 84.0\% for subject 2, and 94.8\% for subject 3. These results demonstrate that the proposed method achieves sufficiently high accuracy for sleep quality classification.

\subsection{Interpretation of Sleep Satisfaction Classification}
In this experiment, we applied TimeSHAP to the trained LSTM model to interpret the classification of sleep satisfaction for a night's sleep. First, we examined the clusters that are important for sleep satisfaction classification across the entire sleep period. We calculated SHAP values for days with satisfied and unsatisfied ratings and compared the magnitude of SHAP values and differences in important clusters. Here, we interpreted days with satisfied ratings by focusing on the positive side and days with unsatisfied ratings by focusing on the negative side of the SHAP values.

\begin{figure}[t]
  \centering
  \includegraphics[width=1.0\linewidth]{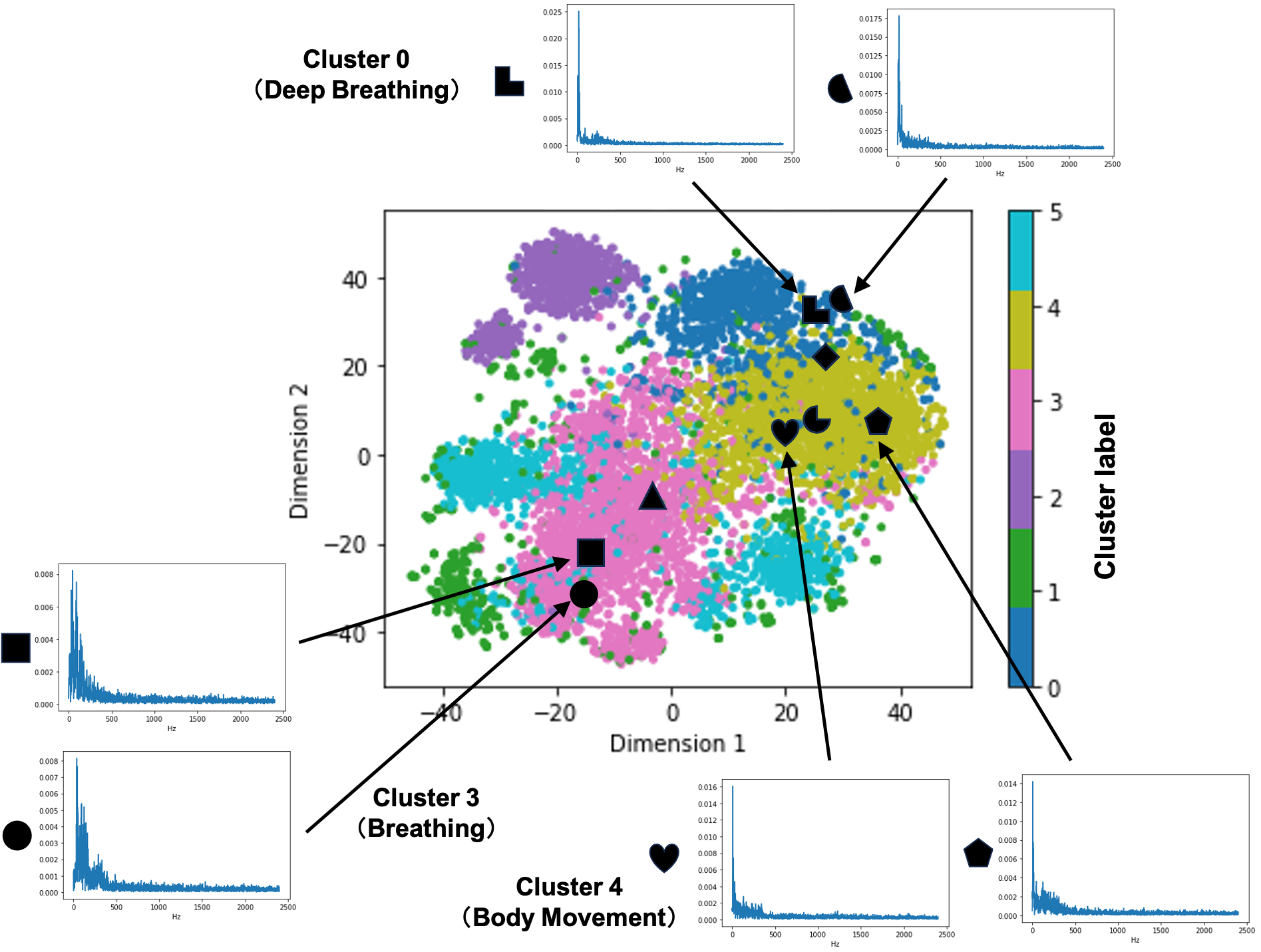}
  \caption{Sleep sound events within important clusters on the latent representation plot by t-SNE (Subject 1)}
  \label{fig:1}
\end{figure}

For each subject, we examined the sleep sound events belonging to the clusters with the largest positive, second-largest positive, and the largest negative SHAP values. While some sounds were impossible to identify upon listening, it was found that in all participants, clusters with large positive SHAP values had a higher proportion of body movement sounds and breathing sounds, indicating their importance on days with high satisfaction. Conversely, clusters with relatively large negative SHAP values had a higher proportion of noise such as deep breathing for Subject 1, and sounds like cars or motorcycles for Subjects 2 and 3, indicating their importance on days with low satisfaction. Furthermore, the ranking of importance for these sleep sound events varied among subjects. As an illustration, Figure \ref{fig:1} shows the correspondence between sleep sound events and the latent space plots for important clusters using t-SNE for Subject 1.

As a representative example, the results of applying TimeSHAP to Subject 1 are shown in Figure \ref{fig:main}. Results for Subjects 2 and 3 are provided in the appendix. Here, ``Pruned Events'' represent features from past events that have been pruned in the direction of sequence events to reduce computational complexity. On days with satisfied, Cluster 3 (breathing sounds) stands out with significantly large SHAP values. Conversely, on days with low satisfaction, only Cluster 0 (deep breathing sounds) has a negative value, but the difference in SHAP values between clusters is not substantial. This indicates that for Participant 1, days with high satisfaction are characterized by a high importance of breathing sounds, while days with low satisfaction are characterized by a relatively higher importance of deep breathing sounds. It is conjectured that these deep breathing sounds occur after apnea, followed by a deep breathing.

\begin{figure}[t]
\centering
    \begin{subfigure}{0.6\linewidth}
        \centering
        \includegraphics[width=0.98\linewidth]{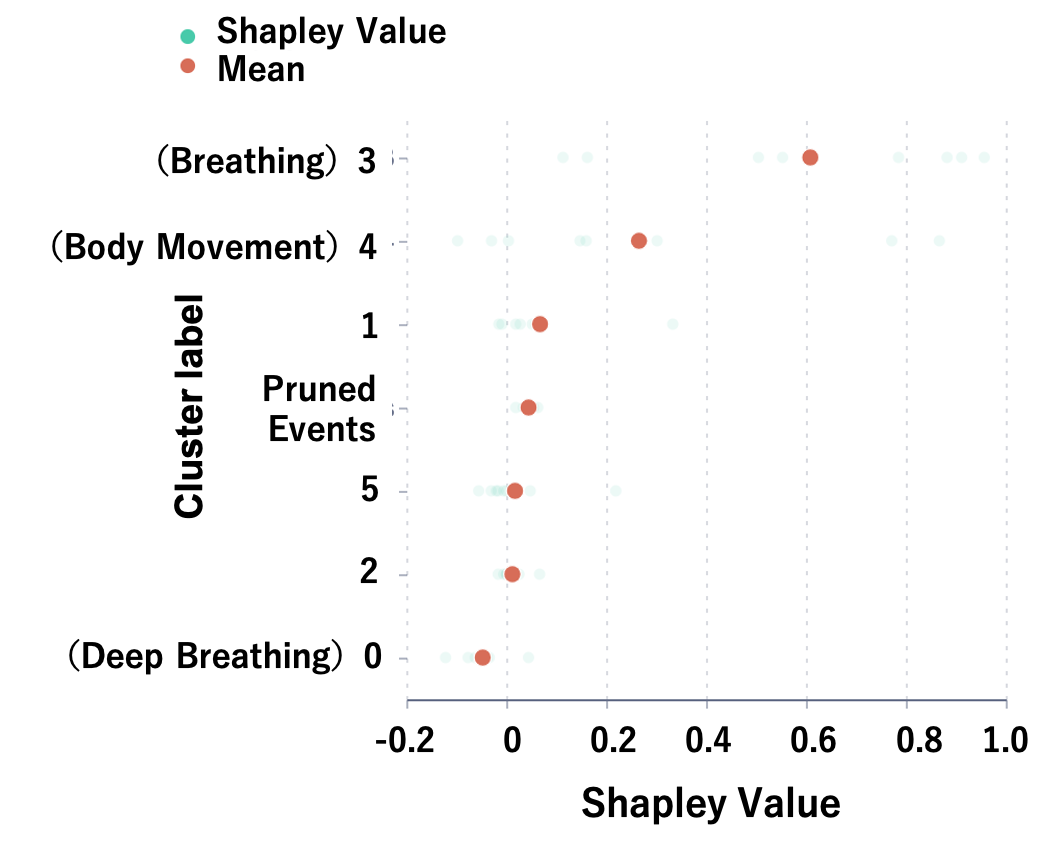}
        \caption{Satisfied days}
    \end{subfigure}
    \begin{subfigure}{0.6\linewidth}
        \centering
        \includegraphics[width=0.98\linewidth]{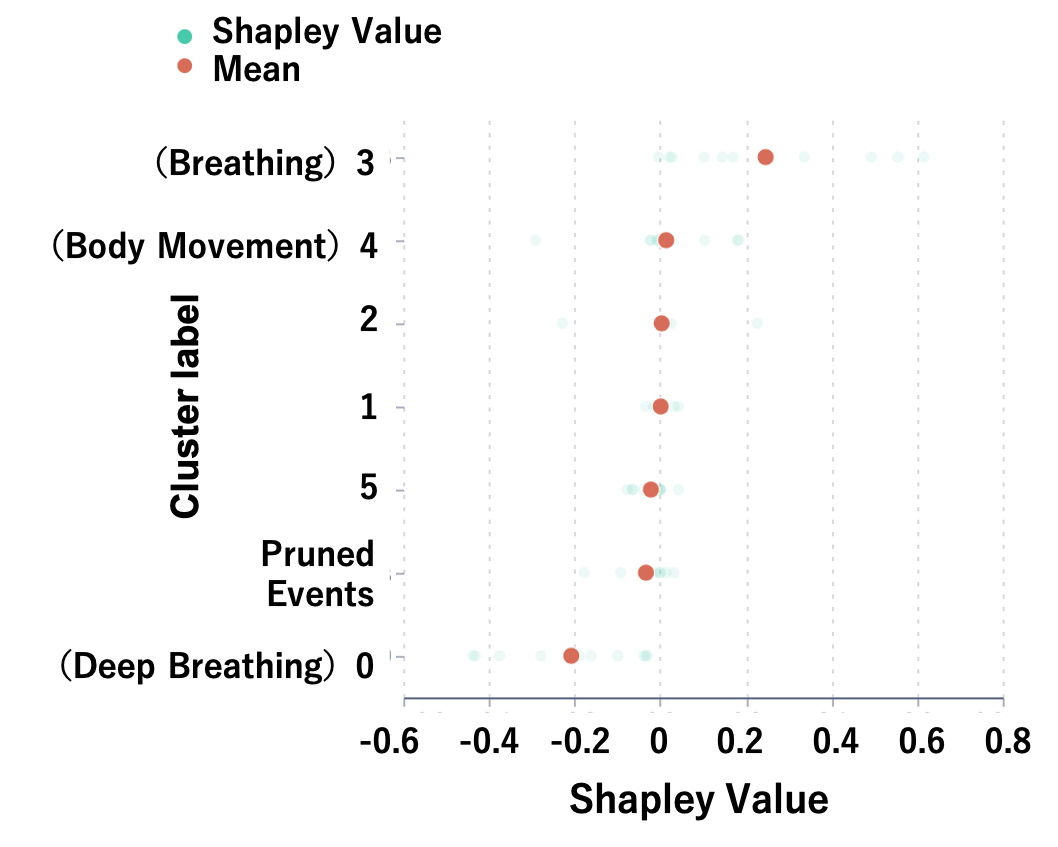}
        \caption{Unsatisfied days}
    \end{subfigure}
    \caption{SHAP values for different satisfactions (Subject 1)}
\label{fig:main}
\end{figure}

Next, we examined the importance of time periods and their respective features for sleep quality assessment, considering the time series aspect. Since TimeSHAP aggregates earlier parts of the sequence into pruned events to reduce computational complexity, applying TimeSHAP to a model trained on data from an entire night of sleep does not allow us to assess the importance of the early stages of sleep. Therefore, we divided the sequence data into three equal parts based on the sequence of events: early, middle, and late stages of sleep, and compared the magnitudes of SHAP values and differences in important clusters. The model to which TimeSHAP was applied was trained on data divided into these three segments according to the sequence of sound events.

\begin{figure}[t]
\centering
    \begin{subfigure}{0.55\linewidth}
        \centering
        \includegraphics[width=0.98\linewidth]{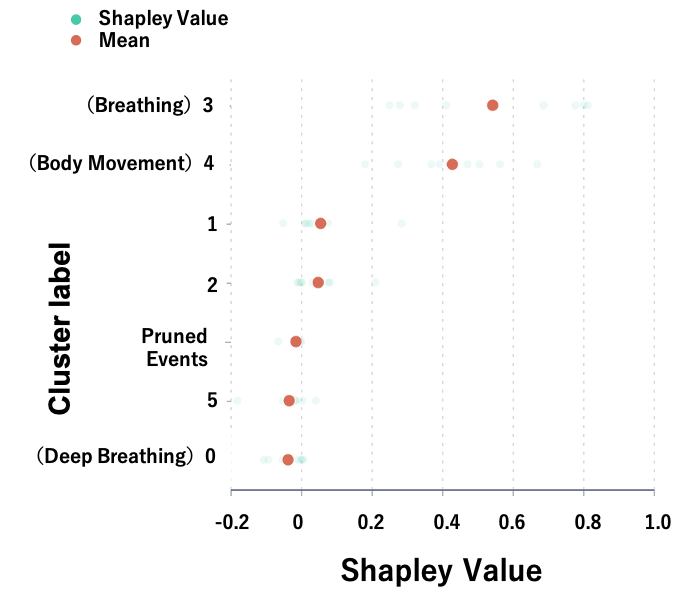}
        \caption{Early}
    \end{subfigure}
    \begin{subfigure}{0.55\linewidth}
        \centering
        \includegraphics[width=0.98\linewidth]{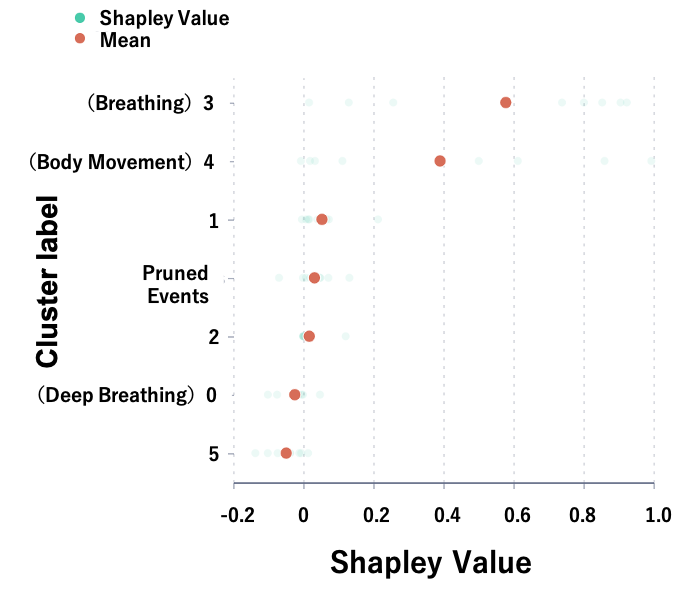}
        \caption{Middle}
    \end{subfigure}
    \begin{subfigure}{0.55\linewidth}
        \centering
        \includegraphics[width=0.98\linewidth]{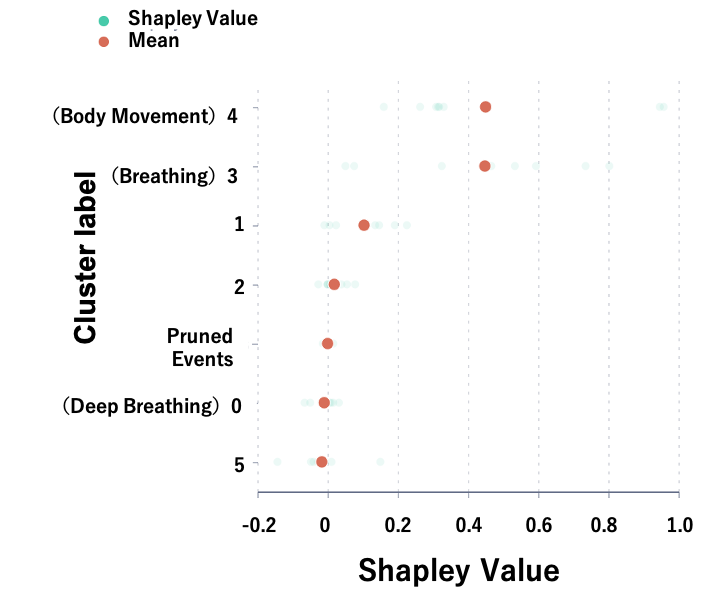}
        \caption{Late}
    \end{subfigure}
    \caption{SHAP values on divided into three equal parts on \textbf{satisfied} days (Subject 1)}
    \label{fig:t1}
\end{figure}

\begin{figure}[t]
\centering
    \begin{subfigure}{0.55\linewidth}
        \centering
        \includegraphics[width=0.98\linewidth]{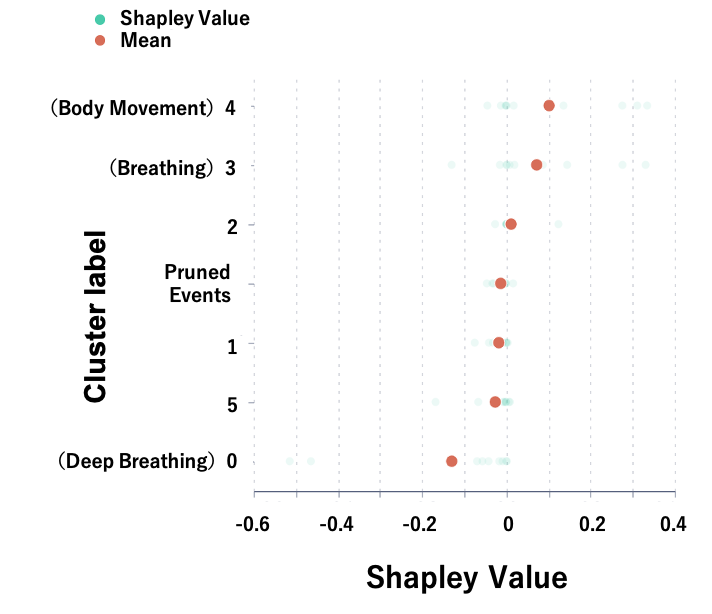}
        \caption{Early}
    \end{subfigure}
    \begin{subfigure}{0.55\linewidth}
        \centering
        \includegraphics[width=0.98\linewidth]{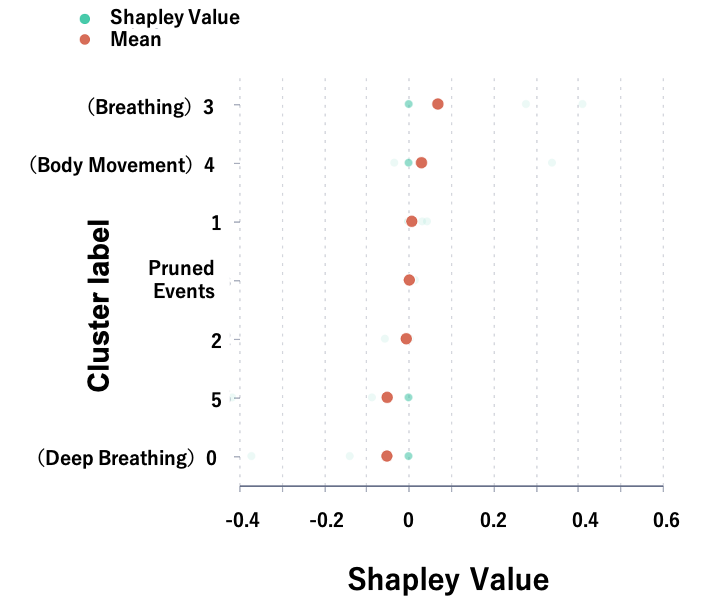}
        \caption{Middle}
    \end{subfigure}
    \begin{subfigure}{0.55\linewidth}
        \centering
        \includegraphics[width=0.98\linewidth]{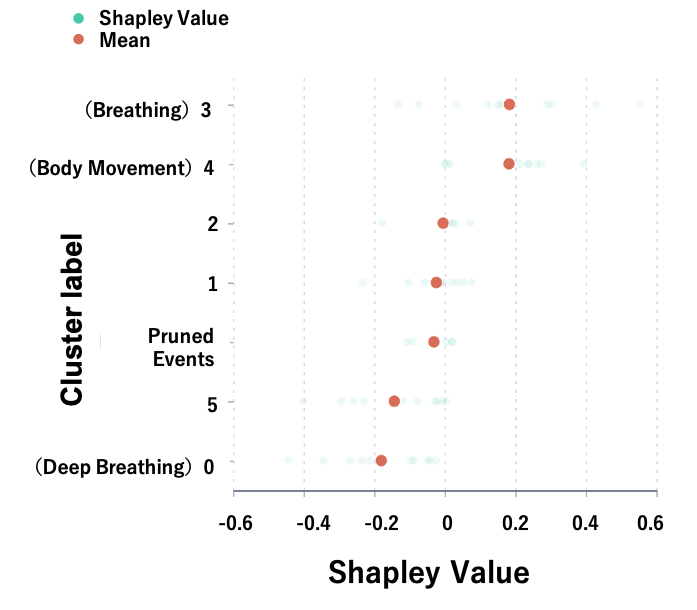}
        \caption{Late}
    \end{subfigure}
    \caption{SHAP values on divided into three equal parts on \textbf{unsatisfied} days (Subject 1)}
    \label{fig:t0}
\end{figure}

The experiment results revealed that the important time periods and their respective features vary among subjects. This suggests that the proposed method can analyze individual sleep characteristics. As an example, the SHAP values for each of the three segments of the sequence for Subject 1 are shown in Figures \ref{fig:t1} and \ref{fig:t0}. The sleep issues observed on days with unsatisfied for each subject are as follows:

\begin{itemize}
\item For Subject 1, the SHAP values of Cluster 0 (large breathing sounds) are relatively large negative values in the early and late segments. However, the differences in SHAP values between clusters in the middle segment are smaller compared to the early and late segments, and their absolute values are also smaller. This indicates that the presence of large breathing sounds in the early and late segments is problematic, suggesting a possibility of sleep apnea syndrome (including pre-syndrome).
\item For Subject 2, the SHAP values for noise are significantly large negative values in all time periods, suggesting that noise is likely to disrupt sleep.
\item For Subject 3, the SHAP values for noise are significantly large negative values in the early and late segments, similar to Subject 2, indicating that noise is likely to disrupt sleep.
\end{itemize}

\clearpage

\subsection{Comparison with the Conventional Method}
We compared the accuracy of sleep satisfaction classification between the proposed method and the conventional method, which used VAE for extracting latent representations of sleep sound events and directly input to LSTM.

To ensure approximately equal computational time per epoch between the proposed and conventional methods, we adjusted the data augmentation factor for the conventional method. We then compared the accuracy of sleep quality assessment for each method. The data augmentation rate was set to 2,000x for subjects 1 and 2 and 5,000x for subject 3, for both the proposed method and the conventional method. Additionally, we conducted a one-sided t-test with a significance level of 5\% to confirm whether there was a significant difference in the mean values between the proposed and conventional methods.

The experimental results are presented in Table \ref{table:pre}. For the conventional method, the latent space dimension of VAE and the multiplying factor of data augmentation were (100, 120) for Subject 1, (20, 300) for Subject 2, and (150, 200) for Subject 3, respectively. The values in the table for the conventional method were obtained using the optimal latent space dimension of VAE. For subjects 1 and 3, the proposed method achieved a statistically significantly higher mean accuracy compared to the conventional method, while for subject 2, the results were comparable. Therefore, it can be concluded that the proposed method achieves at least the same level of accuracy as the conventional method when compared.

\begin{table}[t]
 \caption{Comparison between the proposed method and the conventional method}
 \label{table:pre}
 \centering
  \begin{tabular}{c||ccc}
   \hline
   ID & Proposed & Conventional & statistical significance\\
   \hline \hline
   1 &\textbf{0.903 $\pm$ 0.040} & 0.833 $\pm$ 0.023 &\checkmark\\
   2 & 0.840 $\pm$ 0.073 & 0.840 $\pm$ 0.049 & \\
   3 & \textbf{0.948 $\pm$ 0.050} & 0.916 $\pm$ 0.015 & \checkmark\\
   \hline
  \end{tabular}
\end{table}

\section{Conclusion}
In this study, we proposed a novel method for sleep quality classification over a single night using VAE and LSTM by introducing clustering of sleep sound events and using TimeSHAP to identify the types of sound events involved in the classification as explanations. Through an experiment with three subjects conducted in a home environment over approximately 20 days, we demonstrated that there is a significant difference in the importance of clusters depending on sleep satisfaction, and we analyzed each subject’s unique sleep characteristics, suggesting possible areas for improvement. Additionally, we proposed a method that uses the probability of belonging to each cluster as a representation of sound events. By combining this with data augmentation of event sequences, we observed a trend toward improved classification accuracy compared to the conventional method in the experiments with the three subjects.

The future challenges are as follows: The first challenge is that the number of subjects in this study was limited to three, so it will be necessary to verify the generalizability of the method with a larger number of subjects.

The second challenge is that the optimal dimensionality of the VAE and the number of clusters vary significantly between individuals. When deploying applications for sleep evaluation using the proposed method, an efficient tuning approach will be needed.

The third challenge is the manual labeling of sleep sound events. Since there are individual differences in sleep sound events, even for the same phenomena, it is necessary to label sound events for each person. Manual labeling would be difficult when scaling up to a large number of users. Therefore, it will be necessary to streamline the labeling process by training event classifiers or using transfer learning.

\appendix
\section{Results of Interpretation of Sleep Satisfaction Classification}
\begin{figure}[t]
  \centering
  \includegraphics[width=1.0\linewidth]{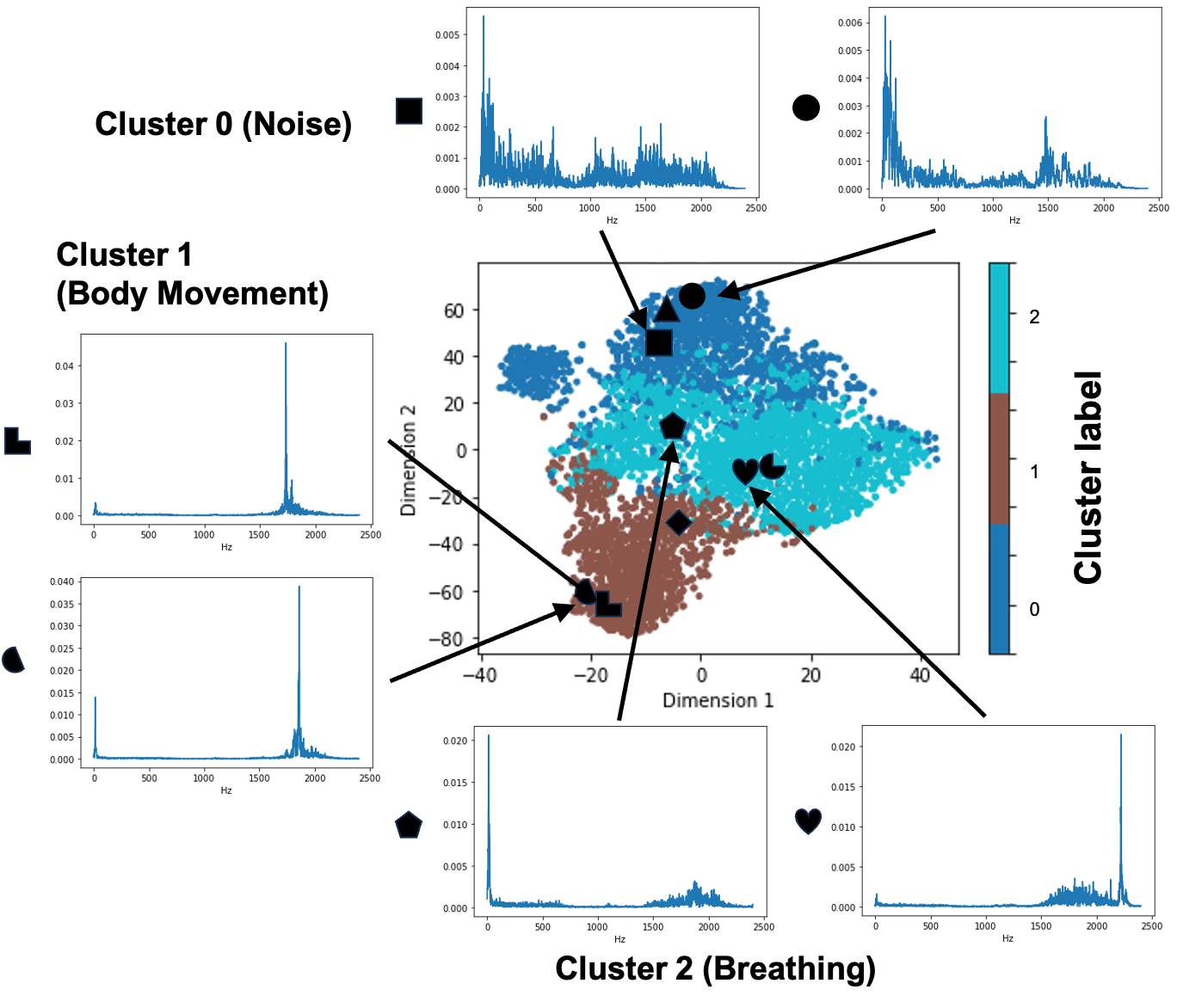}
  \caption{Sleep sound events within important clusters on the latent representation plot by t-SNE (\textbf{Subject 2})}
\end{figure}

\begin{figure}[t]
\centering
    \begin{subfigure}{0.8\linewidth}
        \centering
        \includegraphics[width=0.98\linewidth]{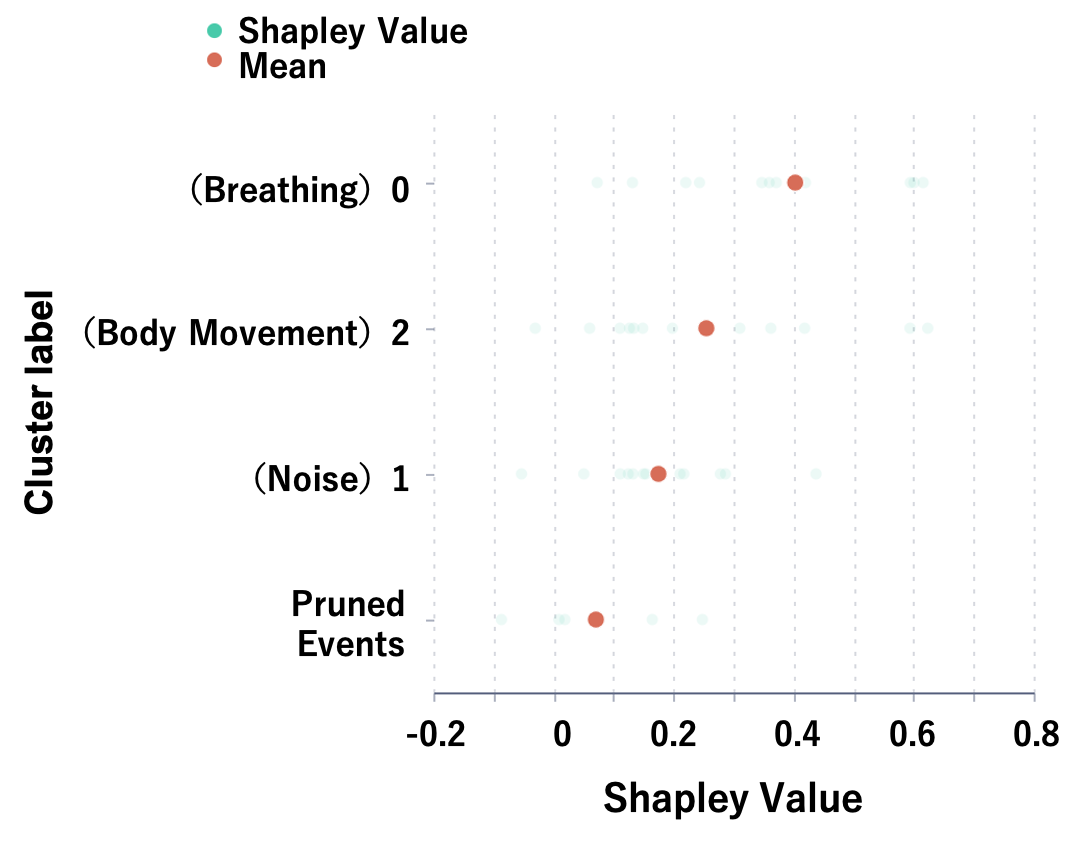}
        \caption{Satisfied days}
    \end{subfigure}
    \begin{subfigure}{0.8\linewidth}
        \centering
        \includegraphics[width=0.98\linewidth]{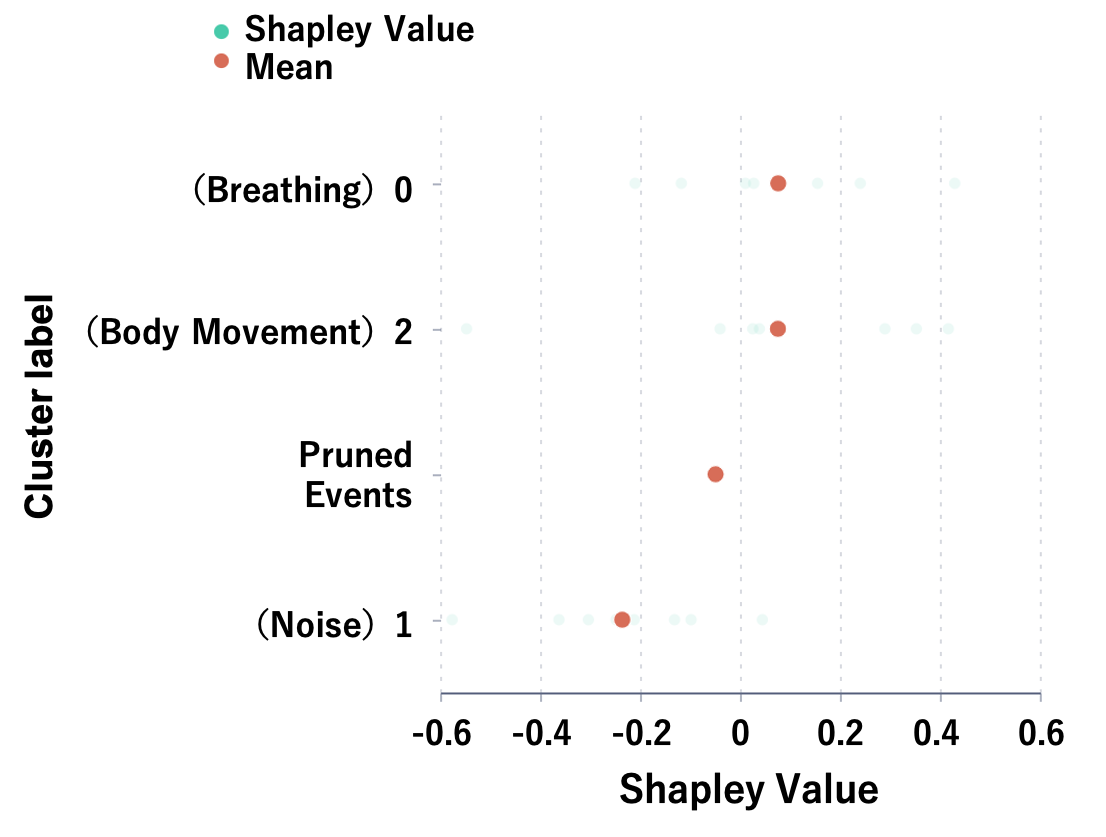}
        \caption{Unsatisfied days}
    \end{subfigure}
    \caption{SHAP values for different satisfactions (\textbf{Subject 2})}
\end{figure}

\begin{figure}[t]
\centering
    \begin{subfigure}{0.5\linewidth}
        \centering
        \includegraphics[width=0.98\linewidth]{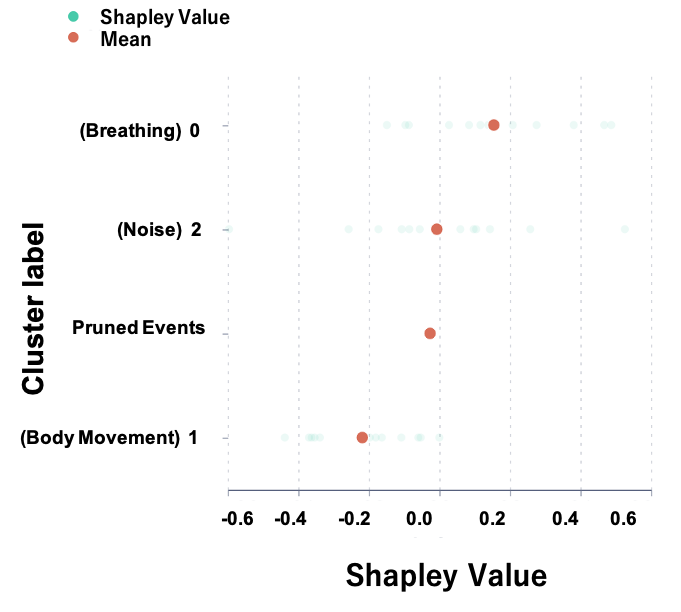}
        \caption{Early}
    \end{subfigure}
    \begin{subfigure}{0.5\linewidth}
        \centering
        \includegraphics[width=0.98\linewidth]{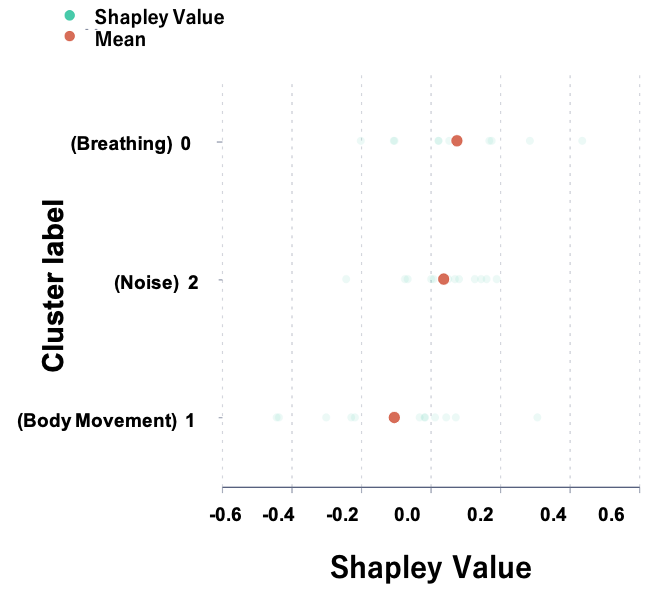}
        \caption{Middle}
    \end{subfigure}
    \begin{subfigure}{0.5\linewidth}
        \centering
        \includegraphics[width=0.98\linewidth]{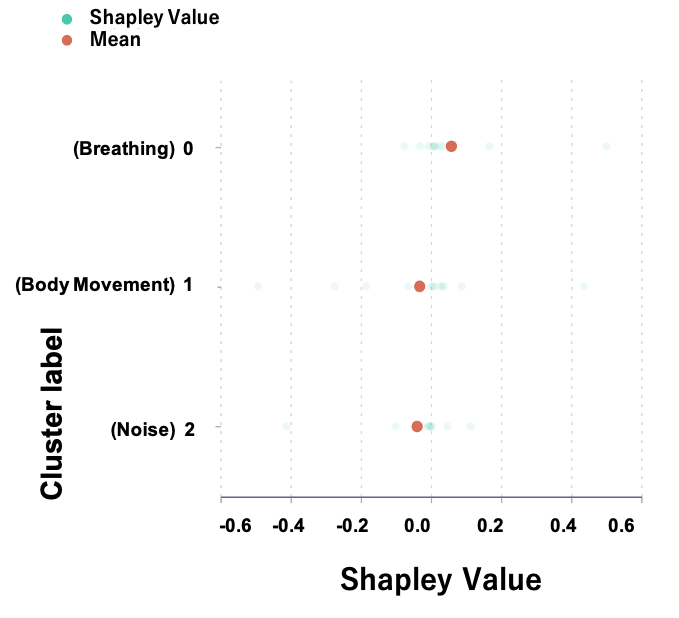}
        \caption{Late}
    \end{subfigure}
    \caption{SHAP values on divided into three equal parts on \textbf{satisfied} days (\textbf{Subject 2})}
\end{figure}

\begin{figure}[t]
\centering
    \begin{subfigure}{0.5\linewidth}
        \centering
        \includegraphics[width=0.98\linewidth]{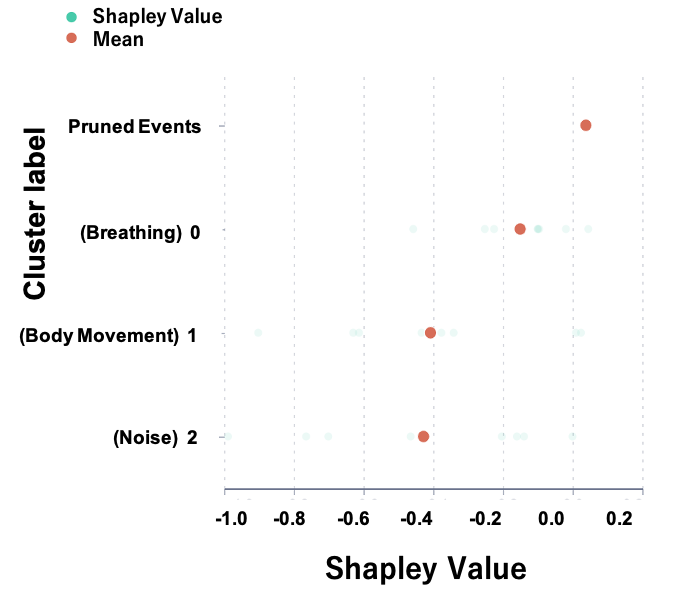}
        \caption{Early}
    \end{subfigure}
    \begin{subfigure}{0.5\linewidth}
        \centering
        \includegraphics[width=0.98\linewidth]{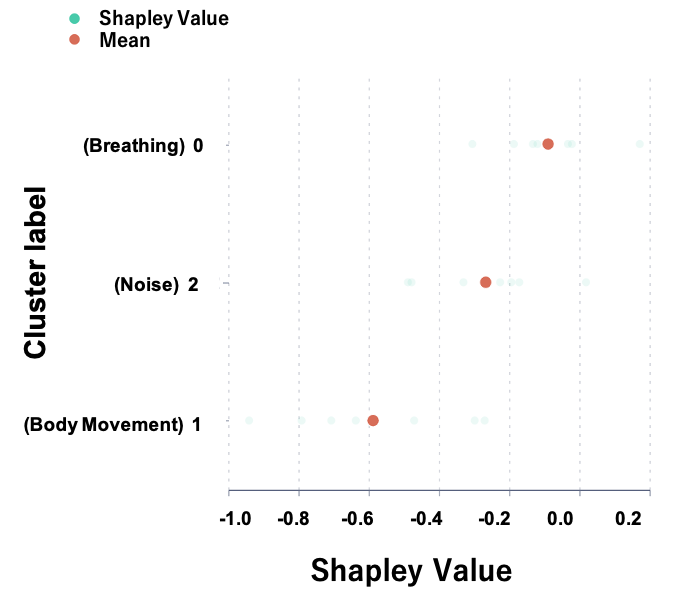}
        \caption{Middle}
    \end{subfigure}
    \begin{subfigure}{0.5\linewidth}
        \centering
        \includegraphics[width=0.98\linewidth]{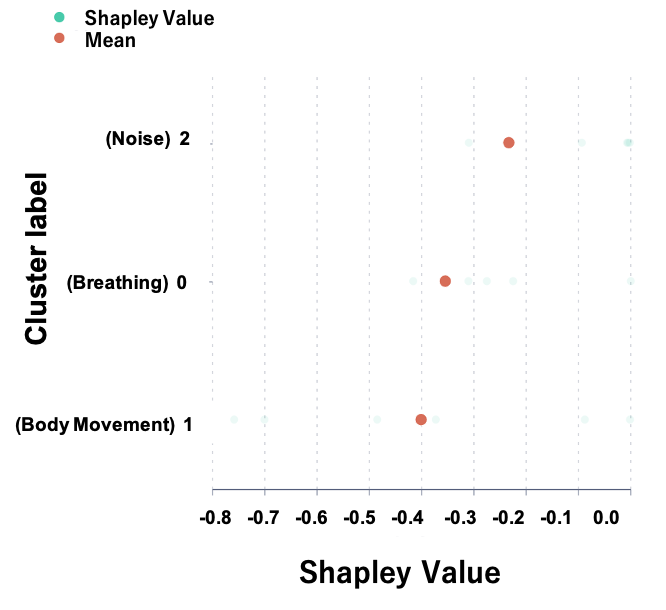}
        \caption{Late}
    \end{subfigure}
    \caption{SHAP values on divided into three equal parts on \textbf{unsatisfied} days (\textbf{Subject 2})}
\end{figure}

\begin{figure}[t]
  \centering
  \includegraphics[width=1.0\linewidth]{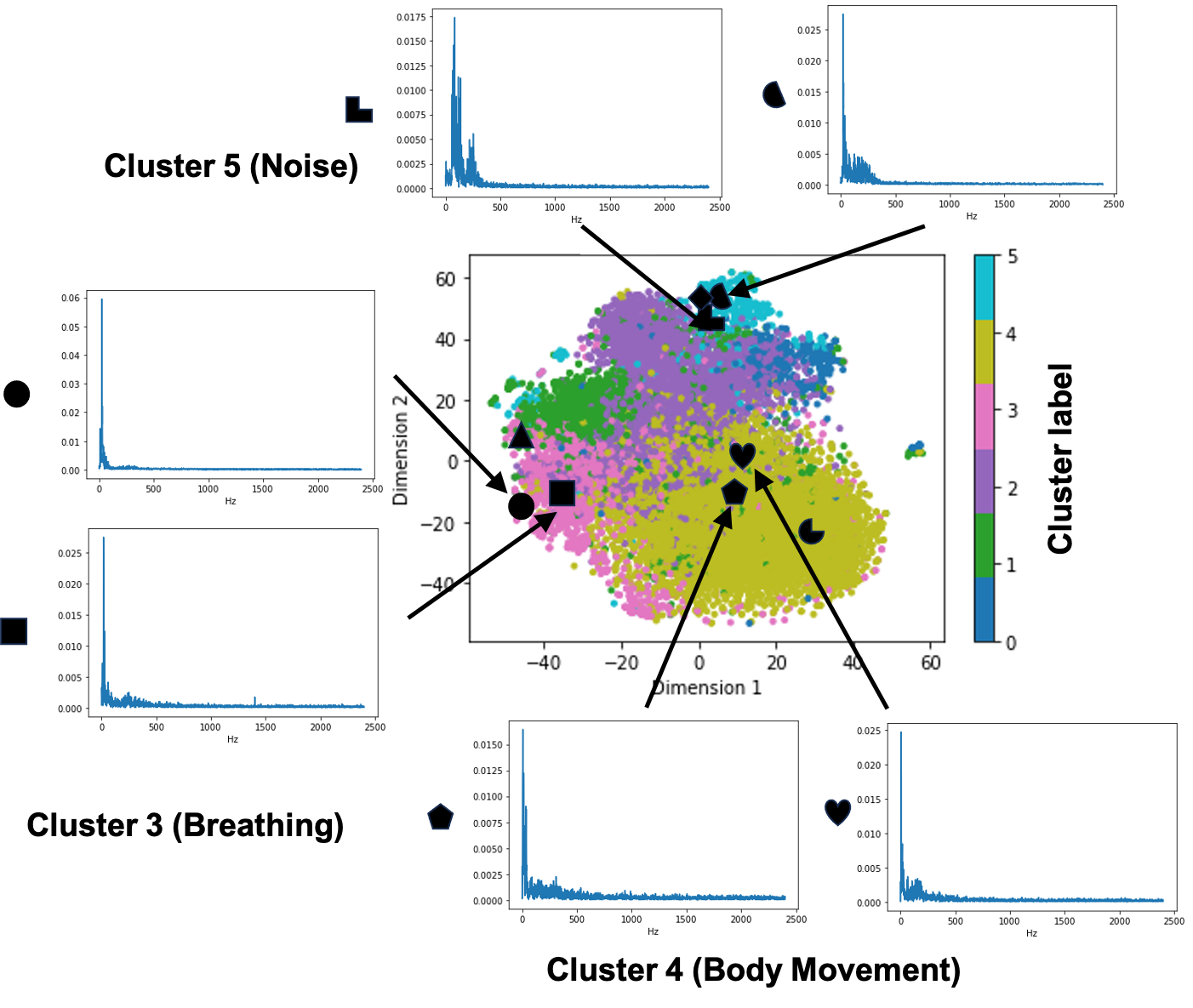}
  \caption{Sleep sound events within important clusters on the latent representation plot by t-SNE (\textbf{Subject 3})}
\end{figure}

\begin{figure}[t]
\centering
    \begin{subfigure}{0.8\linewidth}
        \centering
        \includegraphics[width=0.98\linewidth]{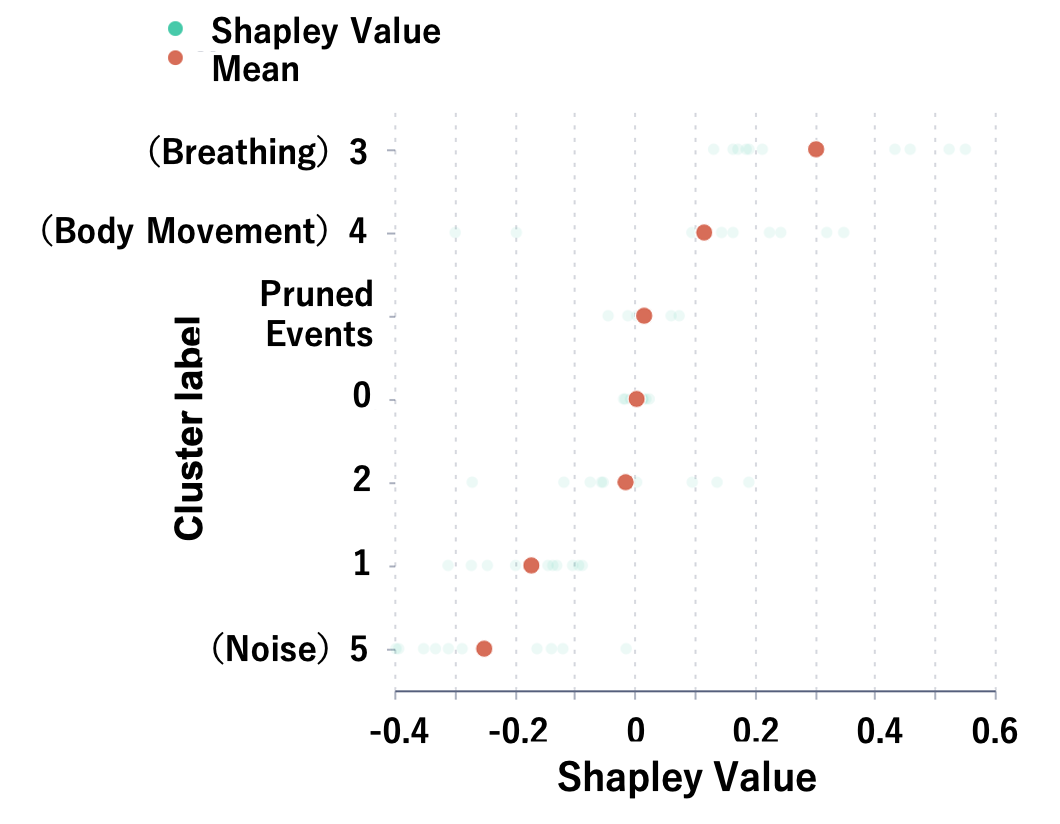}
        \caption{Satisfied days}
    \end{subfigure}
    \begin{subfigure}{0.8\linewidth}
        \centering
        \includegraphics[width=0.98\linewidth]{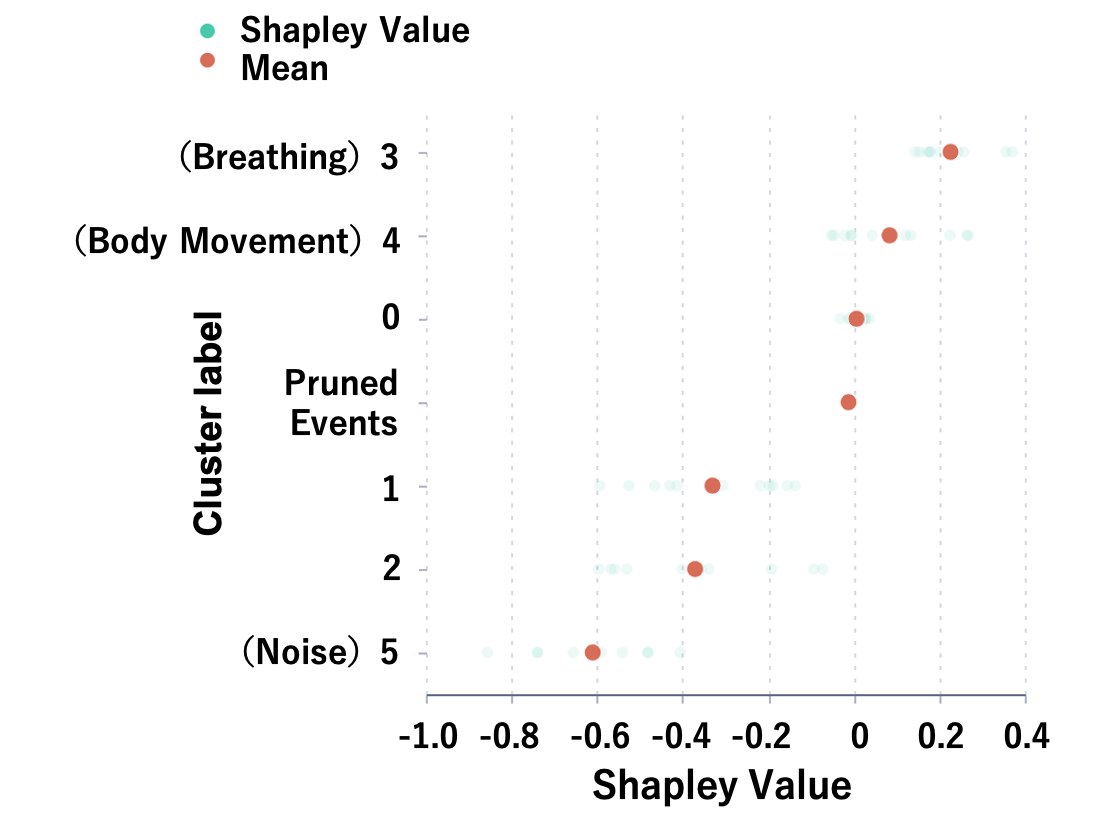}
        \caption{Unsatisfied days}
    \end{subfigure}
    \caption{SHAP values for different satisfactions (\textbf{Subject 3})}
\end{figure}

\begin{figure}[t]
\centering
    \begin{subfigure}{0.5\linewidth}
        \centering
        \includegraphics[width=0.98\linewidth]{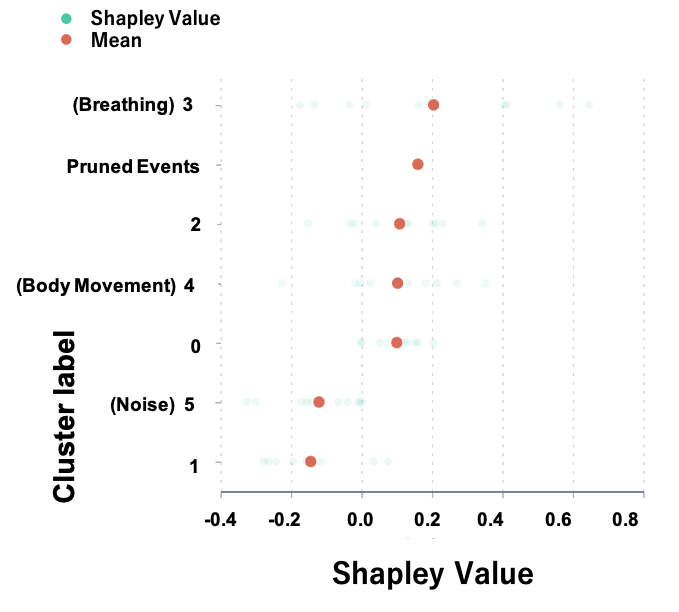}
        \caption{Early}
    \end{subfigure}
    \begin{subfigure}{0.5\linewidth}
        \centering
        \includegraphics[width=0.98\linewidth]{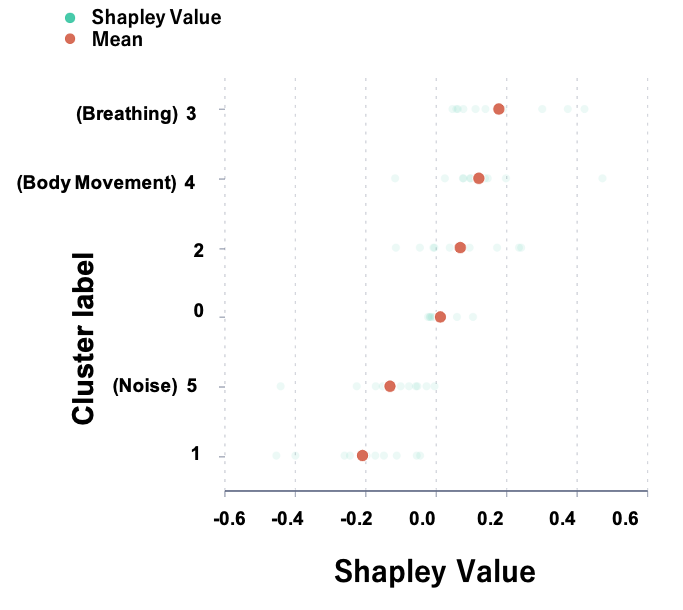}
        \caption{Middle}
    \end{subfigure}
    \begin{subfigure}{0.5\linewidth}
        \centering
        \includegraphics[width=0.98\linewidth]{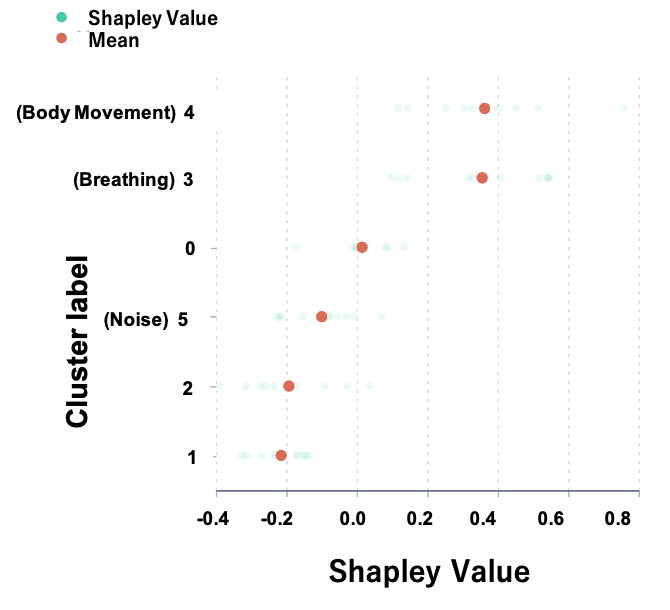}
        \caption{Late}
    \end{subfigure}
    \caption{SHAP values on divided into three equal parts on \textbf{satisfied} days (\textbf{Subject 3})}
\end{figure}

\begin{figure}[t]
\centering
    \begin{subfigure}{0.5\linewidth}
        \centering
        \includegraphics[width=0.98\linewidth]{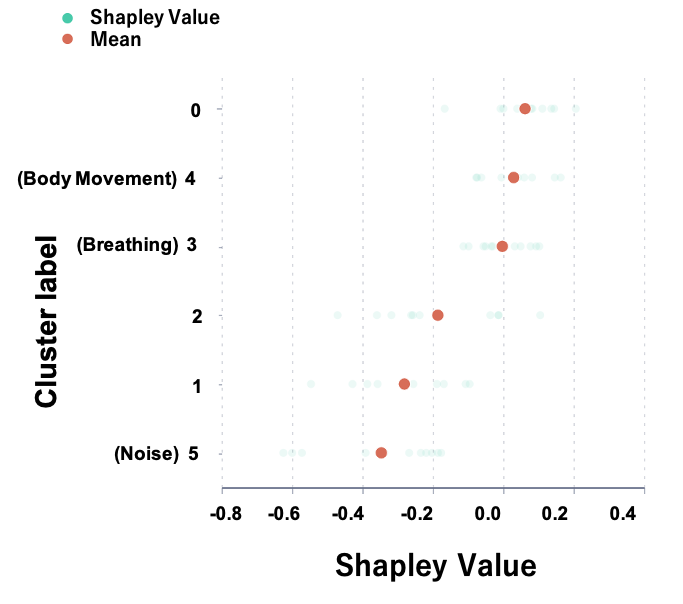}
        \caption{Early}
    \end{subfigure}
    \begin{subfigure}{0.5\linewidth}
        \centering
        \includegraphics[width=0.98\linewidth]{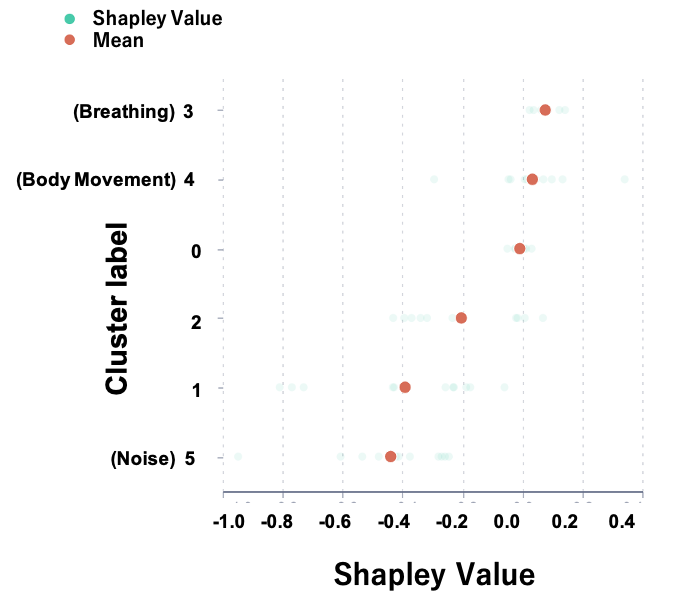}
        \caption{Middle}
    \end{subfigure}
    \begin{subfigure}{0.5\linewidth}
        \centering
        \includegraphics[width=0.98\linewidth]{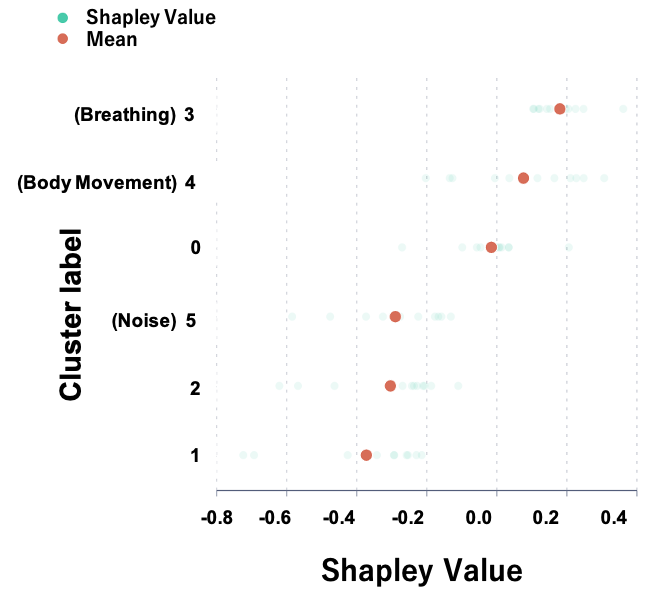}
        \caption{Late}
    \end{subfigure}
    \caption{SHAP values on divided into three equal parts on \textbf{unsatisfied} days (\textbf{Subject 3})}
\end{figure}

\end{document}